\documentclass{article}
\usepackage{spconf,amsmath,epsfig, subfigure, graphicx}
\usepackage{xcolor}

\title{BrightEarth Roads: Towards fully automatic road network extraction from satellite imagery}

\name{Liuyun Duan\textsuperscript{a,}, Willard Mapurisa\textsuperscript{b}, Maxime Leras\textsuperscript{a}, Leigh Lotter\textsuperscript{b}, Yuliya Tarabalka\textsuperscript{a} } 
\address{\textsuperscript{a} LuxCarta Technology, Mouans Sartoux, France 
\textsuperscript{b} LuxCarta South Africa, Cape Town, South Africa}

\begin{document}

\maketitle
\begin{abstract}
The modern road network topology comprises intricately designed structures that introduce complexity when automatically reconstructing road networks. While open resources like OpenStreetMap (OSM) offer road networks with well-defined topology, they may not always be up to date worldwide. In this paper, we propose a fully automated pipeline for extracting road networks from very-high-resolution (VHR) satellite imagery. Our approach directly generates road line-strings that are seamlessly connected and precisely positioned. The process involves three key modules: a CNN-based neural network for road segmentation, a graph optimization algorithm to convert road predictions into vector line-strings, and a machine learning model for classifying road materials. Compared to OSM data, our results demonstrate significant potential for providing the latest road layouts and precise positions of road segments.
\end{abstract}

\section{Introduction}
\label{sec:intro}
With the development of internet and electronic information technologies, the scale of the earth is no longer as immense as before in our imagination or the view of telescopes. We stay ambitious to duplicate the world into a virtual digital twin. Buildings and roads are two of the top interesting man-made object classes that are indispensable to extract. Applications such as smart cities, intelligent navigation, and simulation require large-scale high accuracy road networks with rich attributes to conduct realistic experiences. The precision of road axis positions, correctness of topology and geometry, and attributes such as number of lanes, width, materials become critically important for most applications. 
Thanks to the continuously intensive developments of remote sensing image processing \cite{modelbased95,eval97} and deep learning technologies, fully automatic road extraction becomes more and more promising \cite{CHEN2022,Lian20review}. Recent road extraction works based on deep learning can be sorted into two major groups: pixel-based road segmentation with a post-processing module to convert the raster prediction to road axis in line-strings \cite{resunet, Vnet20, Zhou18-dlinknet}; graph reconstruction directly from images by iterative prediction of next connecting point or by detecting interesting points of road topology with/without a segmentation guidance \cite{TDroad, Tan20_vecroad, gaetan_2022_CVPR,Mattyus_2017_ICCV}. However, most of these existing methods often do not handle relatively simple scenarios with sparse intersections such as in residential areas. Incompleteness and broken connections are the most common problems when applying these methods to large-scale road network extraction. In this paper we propose an improved CNN-based road segmentation method and a graph optimization algorithm to generate road networks with mathematically explainable controls of the quality of the geometry and the topology. The precise road geometry and reliable topology also aid in the classification of road materials.

We summarize our main contributions as following:\\
1) Effective and efficient automatic road extraction pipeline that leverages both the advantages of high performance deep learning segmentation and robust and controllable remote sensing image processing technology.\\
2) New road segmentation architecture that improves the prediction quality by leveraging multiple learning tasks in one framework.\\
3) New road network reconstruction method based on graph optimization that generates OSM-like road structures with geometrical compactness and regularity.\\
4) High performance road material classification module by applying a light machine learning model with a small amount of training data, which is supported by the accurate road network vector layer generated by the previous two points.

We describe the technical details of the proposed pipeline in the following three sections: road segmentation, road network reconstruction and road material classification.

\begin{figure}
\centering
\includegraphics[width=0.7\linewidth]{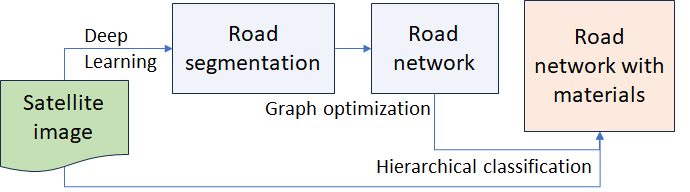}
\caption{The proposed road extraction pipeline from a single input satellite image.}
\label{fig:pipeline}
\end{figure}

\section{Road Segmentation}
\label{sec:seg}
The extraction of road semantics is subject to several challenges such as occlusion from buildings, trees and other roads, abrupt changes in color and materials along the same road and road-like human-made objects. To overcome these issues, the combination of a precise and worldwide dataset, a state of the art neural network design and a multi-class loss is needed.

\subsection{Data Preparation}
Our dataset comprises 130 km² areas of 30cm and 50cm satellite imagery, covering 108 zones worldwide. Each image was manually annotated with polygons of sub-pixel precision Ground Truth to match sidewalks and road limits. The road polygons were rasterized into a three-class mask that represents road interiors, road contours and others. Rasterizing these polygons without considering overlapping roads, as seen in interchanges, was insufficient. We added information to the polygons to determine the relative road levels locally.


\subsection{Neural Network Design}

\begin{figure}
\centering
\includegraphics[width=0.6\linewidth]{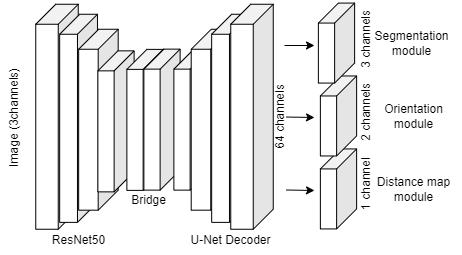}
\caption{The proposed new architecture for road segmentation from a single satellite image.}
\label{fig:architect}
\end{figure}

The architecture of our road model is based on ResNet for the backbone and U-Net for the decoding part \cite{resunet}. We have three modules responsible for the last deconvolutions: the segmentation module that outputs the three classes (road interior, road contour and others); the orientation module that outputs local orientations \cite{girard2021polygonal}; and a distance map module \cite{Dangi_2019}. The orientation module, trained with a smoothing loss, helps to segment obstructed roads in a continuous way. The distance map module benefits both interior and contour segmentation. It is redundant with the contour, but it offers flexibility to define the best position of contours without too much penalty when missing the contour by one pixel. This is in contrast to IoU (Intersection over Union) or cross-entropy losses on the three classes segmentation. Figure \ref{fig:architect} illustrates the idea of our neural network design.

\subsection{Training}

Our model is trained on the dataset with a training/validation ratio of 85/15. The segmentation module is trained using IoU and cross-entropy losses with more weight on contours to balance classes. The orientation module is trained using a differentiable loss on predicted angles so that a predicted orientation of $\pi$ is equivalent to a predicted orientation of 0. The distance map module is trained using L2 norm pixel-wise to penalize greater differences. Examples of our road segmentation are shown in Figure \ref{fig:res}, second column.

\section{Road Network Reconstruction}
\label{sec:graph}
Following the segmentation of the roads, the reconstruction of the road network is performed, where the results of the segmentation are vectorized. The goal of reconstruction of the road network is to reconstruct road vectors that accurately follow the center of the road. This process is composed of five processes; namely road skeletonization, linear road recovery, denoising, road smoothing, and road circle reconstruction. 

 \subsection{Road Skeletonization}
From the segmentation results, a road skeleton \cite{lee1994building, zhang1984fast} is created and vectorized as the first approximation of the road center. A graph is created to represent the vectorized skeleton. The vectorized skeleton is characterized by a stair stepping effect as a result of the raster to vector conversion. 

\subsection{Linear Roads Recovery}
To remove the stair stepping effect, an incremental linear least squares line fitting is performed on the vector skeleton in between road junctions to create linear road vectors and remove stair stepping. A junction is any point where three or more roads intersect. Incremental least squares result in linear road vectors represented by the minimum number of vertices between junctions. 

\subsection{De-Noising}
\label{ssec:denois}
After recovering the initial linear road vectors, a cleaning operation is performed, termed denoising. The cleaning operation removes invalid roads by identifying closed road loops that cover small areas, since small areas do not constitute meaningful road features, with the exception of circles. By identifying loops with areas below a defined threshold in the graph of road vectors, noise and errors are removed. This is achieved by converting the loops into linear road segments. Since some of the loops detected are traffic circles, Hough circle detection \cite{borovicka2003circle} is performed on all detected loops. The raster sections from the segmentation results covered by the detected loops are used as input in verifying the presence of circles. Any loops that cover circles are recorded and used in the final step to reconstruct traffic circles, while all other loops are removed and replaced by linear road segments. This operation removes the noise from the road network. Additionally, dangling roads below a given threshold are filtered out. 

\subsection{Road Smoothing}
One major drawback of raster skeletons is the inability to preserve T-junctions and correct road junction topology. Subsequently, the linear roads recovered in the second step have distorted junctions. After denoising, T-junctions are then reconstructed. This is achieved by imposing a smoothness constraint at the junctions. The constraint is that at intersections, at least two incident vertices to a junction and the junction vertex itself must be collinear. Enforcing this constraint on the vector road graph reconstructs the junctions of the road and results in smooth roads. The defined constraint ensures curves with smoothly changing curvature that is not disrupted by incorrect topology at junctions are reconstructed. Thus long roads with smooth curves are reconstructed. More complex road junctions are reconstructed incrementally.

\subsection{Traffic Circle Reconstruction}
Finally, circle fitting is performed on loops that cover circles that were detected in the second step \ref{ssec:denois}. The original loops are replaced with the detected circles, thereby completing road reconstruction. Furthermore, the sections detected as double lanes are reconstructed by replacing the detected double lane center line with duplicate roads on either side of the double lane detected center line. 

Examples of road networks generated by our method are shown in Figure \ref{fig:res}, third column.

\section{Road material Classification}
\label{sec:material}
A hierarchical classification method is proposed to distinguish the material of generated road networks. Roads are classified initially into two main surface classes: processed and unprocessed. Processed roads refer to those made out of concrete, tar or other man-made, solid surfaces. Unprocessed roads include those made up of dirt, sand, gravel or other naturally occurring materials. 
To allow for better statistical representations to be extracted from the corresponding images, a small buffer of 2m is applied to the line-strings to create polygon segments to collect road material pixels from the input satellite images. We apply two iterations of the Support Vector Machine (SVM) classifier to train the processed and unprocessed road classification based on 148km² (RGB) and 483km² (RGB-Nir) archived data with hand crafted road material classes tagged in the Ground Truth road networks. 

To further distinguish the subclasses: gravel and sand for unprocessed roads, we introduce our BrightEarth Land Use Land Cover (LULC) layer \cite{brightearth22} (generated from the Sentinel-2 global mosaic) to determine the road materials in a semantic context. We search the land cover labels within the radius of 1~km$^2$ of each unprocessed road. If the surrounding is mostly barren land or water areas, then the probability of this road being sand is higher than gravel.

\begin{figure}
\centering
\includegraphics[width=0.8\linewidth]{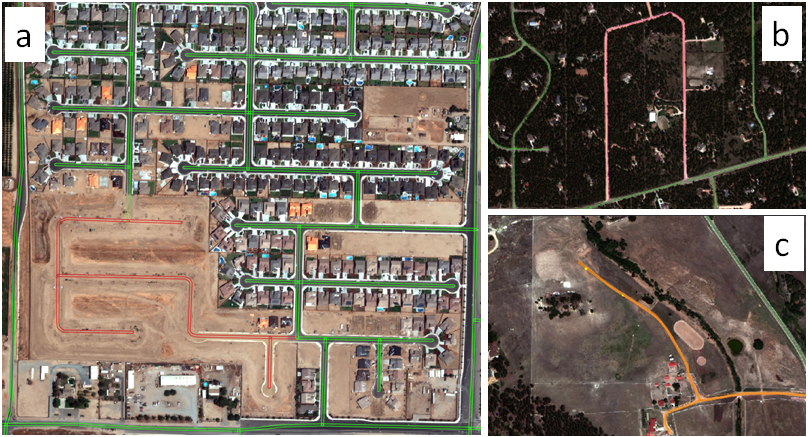}
\caption{Road materials classification examples. Green for processed roads and red for unprocessed roads (a), pink for gravel roads (b) and orange for sand roads (c).}
\label{fig:classif}
\end{figure}

The greatest advantage of the proposed SVM scheme is that the model can be trained with a small amount of data while providing a high generality and robustness to various application scenarios. Our experiments show an overall road precision of 89.99\%, and recall of 84.77\% for road materials classification. Figure \ref{fig:classif} shows visual examples.

\section{Results and Discussion}
\label{sec:res}
We show the performance of our road extraction pipeline by testing on three different styles of cities: Timbuktu in Mali, Amman in Jordan and Aden in Yemen. All images are in 50 cm spatial resolution. The total testing area is 33.4 km\textsuperscript{2}. All the tests are conducted on a local PC with a CPU AMD K19, RAM 16G and GPU NVIDIA GeForce RTX 3060 with 4G memory. We evaluate the quality of our automatically extracted road networks both quantitatively and qualitatively. 

\begin{figure*}[htb]
\centering
\includegraphics[width=0.95\linewidth]{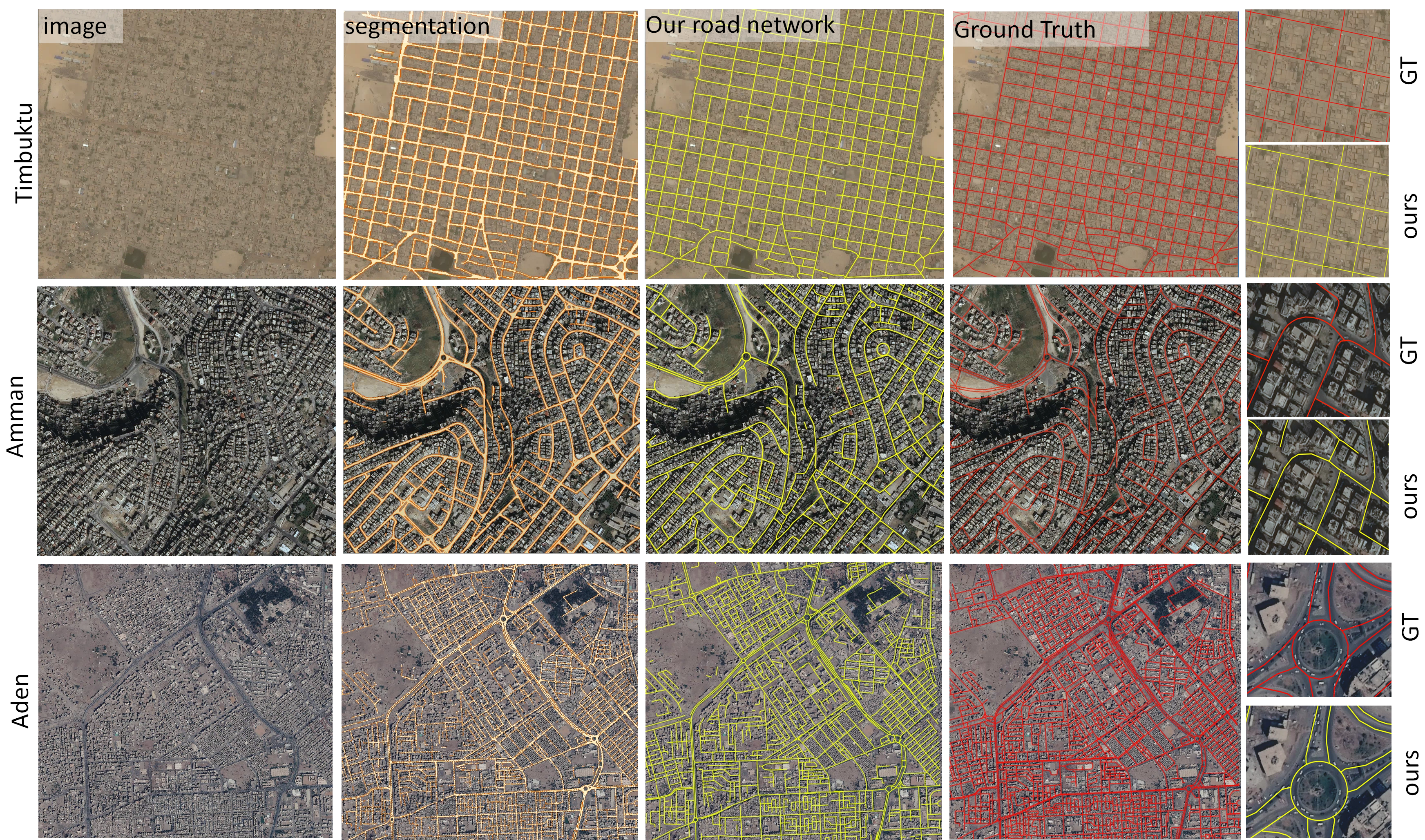}
\caption{Road networks generated by our pipeline over Timbuktu, Amman and Aden (top to bottom rows), with input satellite images, road segmentation, our extracted road networks, the Ground Truth, and crops (left to right columns). In the crops, Ground Truth are marked in red and our results in yellow.}
\label{fig:res}
\end{figure*}

\subsection{Quantitative Statistics}

Our Ground Truths are road polygons that depict along the road contours in the images. To evaluate the quality of our automatically reconstructed road networks, we manually created three sets of road networks as linear Ground Truths: Timbuktu in Mali, Amman in Jordan and Aden in Yemen. We buffer the generated road networks and the Ground Truths with a radius of 2 meters as a relaxation in all the comparisons, denoted as ${\{r_{buff}\}}$ and ${\{gt_{buff}\}}$. We demonstrate the geometrical and topological quality by analyzing the object-wise Precision, Recall, F1score and Hausdorff distance. The accuracy of the positions of our extracted roads is measured by calculating the average Hausdorff distance of the true positive roads and their matched references in the Ground Truth. Here are the definitions of our measure metrics: \\
    \textbf {true positive} for each correctly generated road ${r_{buff}}$, at least 50\% of its area is covered by ${\{gt_{buff}\}}$;\\
    \textbf {false positive} for each redundantly generated road ${r_{buff}}$, less than 50\% of its area is covered by ${\{gt_{buff}\}}$;\\
    \textbf {false negative} for each missing Ground Truth road ${gt_{buff}}$, less than 50\% of its area is covered by ${\{r_{buff}\}}$.

\begin{table}
    \centering
    \begin{tabular}{p{0.43in}|p{0.3in}|p{0.4in}|p{0.3in}|p{0.35in}|p{0.5in}}
    \hline
    \small{city} & \small{GT Length} & \small{Precision} & \small{Recall} & \small{F1score} & \small{Hausdorff distance} \\ 
    \hline
    \small{Timbuktu } & 70.8 &  0.94  & 0.77  & 0.82 & \textbf{0.65}m\\ 
    \hline 
    \small{Amman} & 351.5 &  0.86   & 0.77 & 0.81  & \textbf{0.58}m \\ 
    \hline
     \small{Aden} & 281.7  &  0.87   &  0.68 & 0.74 & \textbf{0.46}m\\  
    \hline
     \small{Average} &   & \textbf{0.87} & \textbf{0.73} & \textbf{0.78} & \textbf{0.54}m\\
    \hline
    \end{tabular}
    \caption{Quantitative evaluations on the three testing areas. \textit{GT length} is the total length of the Ground Truth roads in kilometer. $Average$ values in the last column are weighted averages that take into account of $GT Length$.}
    \label{tab:geometry-eval}
\end{table}

To further demonstrate the quality of our reconstructed road networks, we increase the buffer radius from 2m to 3m, relaxing the constraint of Hausdorff distance. This allows us to include more roads that are topologically matched with the Ground Truth, even if they are slightly shifted away from the centerline of roads. The average Precision, Recall, F1score are \textbf{0.93}, \textbf{0.79}, \textbf{0.85}, and the average Hausdorff distance is \textbf{0.86}m. 

\subsection{Qualitative Results}
Figure \ref{fig:res} illustrates the visualization quality of our results, including input images, road segmentation, and extracted road networks. The crops demonstrate the regularity of the geometry and the potential of our method to preserve the correct topology.

\section{Conclusions and Perspectives}
\label{sec:conc}
We propose a practical fully automatic pipeline for extracting road networks from VHR satellite images. Our pipeline provides geometrical regularity, topological correctness, and material classes. The neural network model we proposed labels road pixels with precise contours, enabling the proposed graph optimization method to generate clean road networks. Our method offers a practical and reliable solution for applications such as simulation that require strictly operational road networks. We aim to enhance the completeness of our road segmentation and continue exploring AI-aided solutions to tackle challenging scenarios, such as hyper-modern areas with elevated roads and bridges that are intertwined.

\clearpage
\bibliographystyle{IEEEbib}
\bibliography{strings,refs}

\end{document}